\documentclass[12pt]{article}
\usepackage{subcaption}
\usepackage[margin=1in]{geometry} 
\usepackage{amsmath,amsthm,amssymb}
\usepackage{graphicx}

\usepackage{natbib}
\usepackage{hyperref}
\usepackage{titling}
\begin{document}
 
%
%
 
\title{Teaching AI to Remember: Insights from Brain-Inspired Replay in Continual Learning} 
\author{Jina Kim\\
School of Computing, KAIST\\
}
\date{}
\maketitle

\vspace{-3em}

\begin{abstract}
Artificial neural networks (ANNs) continue to face challenges in continual learning, particularly due to catastrophic forgetting, the loss of previously learned knowledge when acquiring new tasks. Inspired by memory consolidation in the human brain, we investigate the internal replay mechanism proposed by~\citep{brain_inspired_replay1}, which reactivates latent representations of prior experiences during learning. As internal replay was identified as the most influential component among the brain-inspired mechanisms in their framework, it serves as the central focus of our in-depth investigation. Using the CIFAR-100 dataset in a class-incremental setting, we evaluate the effectiveness of internal replay, both in isolation and in combination with Synaptic Intelligence (SI). Our experiments show that internal replay significantly mitigates forgetting, especially when paired with SI, but at the cost of reduced initial task accuracy, highlighting a trade-off between memory stability and learning plasticity. Further analyses using log-likelihood distributions, reconstruction errors, silhouette scores, and UMAP projections reveal that internal replay increases representational overlap in latent space, potentially limiting task-specific differentiation. These results underscore the limitations of current brain-inspired methods and suggest future directions for balancing retention and adaptability in continual learning systems.

\end{abstract}

\section{Introduction}
\hspace{0.5cm} Despite significant advancements in deep learning, artificial neural networks (ANNs) still suffer from catastrophic forgetting in continual learning, where training on new tasks causes them to easily forget previously learned information. In contrast, the human brain retains diverse information through declarative and nondeclarative memory systems (\citep[Figure 24.1, p.~838]{neuroscience_textbook}), storing it in either short-term or long-term memory. A key factor that protects humans from drastic forgetting is thought to be the reactivation of neural activity patterns representing previous experiences—referred to as memory replay (\cite{brain_memory_1, brain_memory_2, brain_memory_3, brain_memory_4}).

To address catastrophic forgetting in ANNs, previous works have attempted to mimic the brain's memory replay mechanism. Notably, studies such as \cite{brain_inspired_replay1, brain_inspired_replay2, brain_inspired_replay3} have demonstrated that brain-inspired mechanisms can help retain performance during continual learning in AI. Motivated by these findings, we aim to draw inspiration from the brain to develop mechanisms for long-term memory in AI. Specifically, we focus on analyzing the impact of brain-inspired components on AI performance and providing insights to guide future research directions.



\section{Brain-inspired replay for continual learning with artificial neural network}
\hspace{0.5cm} 
To resolve the catastrophic forgetting in ANN, \cite{brain_inspired_replay1} introduces the brain-inspired generative replay(GR) model, which further improves the straightforward generative replay(GR) model. The straightforward GR has been introduced as one solution for mimicking the memory replay in the brain, by generating the replayed data with a generative model learned from past observation and integrated with the new data by the main model. The generator that produces the replayed data is considered the hippocampus, while the main model, which produces the final output from the integrated new data and replayed data, is considered the cortex. Although this model performs well on class-IL scenarios unlike other baselines, it still isn't an optimal solution because of the following reasons. First of all, it is hard to scale up to more challenging problems, and during class IL, it suffers from managing complex inputs. Also, although it tries to mimic the interaction between the cortex and hippocampus, this raises the question of how replay could underlie memory consolidation in the brain. Therefore, \cite[Fig. 7, ~p. 6, 11]{brain_inspired_replay1} proposes a brain-inspired GR model that contains the following brain-inspired components: (1) replay through feedback (2) conditional replay (3) gating based on internal context (4) internal replay (5) distillation. A detailed explanation of each component can be found in \ref{appendix:brain_replay_description}.




\section{Further Analysis}
\label{further_analysis}
\hspace{0.5cm} Although \cite{brain_inspired_replay1} has shown impressive performance(\citep[Fig.~5, 6]{brain_inspired_replay1}) and highlighted the importance of each component in the model(\citep[Fig.~8]{brain_inspired_replay1}), the actual test accuracy is relatively low, ranging around 0.2(\citep[Fig.~8c]{brain_inspired_replay1}). In addition, there are still unresolved questions regarding how and why these brain-inspired factors contribute to mitigating catastrophic forgetting.
Therefore, we provide a further analysis to understand the model's performance and investigate its actual impact as well as potential improvement points. Motivated by the observation that internal replay is the most critical component of the model (\citep[Fig.~8c]{brain_inspired_replay1}), we focus on internal replay and seek to answer the following questions: (1) How do the retention ratio, forgetting score, initial(final) test accuracy differ for each task? (2) How does internal replay influence the log-likelihood distribution and reconstruction error distribution? (3) How does internal replay impact hidden layer activations?

\subsection{Experiment Settings}
\hspace{0.5cm} 
We compare model with internal replay(w/ IR) and model without internal replay(w/o IR), and since brain-inspired replay with synaptic intelligence (BIR + SI) gives the highest accuracy(\citep[Fig.~6c]{brain_inspired_replay1}), we conduct experiments both on brain-inspired replay(BIR) and brain-inspired replay with synaptic intelligence (BIR + SI). Accordingly, the models used in our experiments are as follows: BIR(w/ IR), BIR(w/o IR), BIR + SI(w/ IR), BIR + SI(w/o IR). We train and test on CIFAR-100 in class-IL scenario, since the models mostly suffer on class-IL than task-IL(\citep[Fig.~6b, 6c]{brain_inspired_replay1}) and since BIR(w/o IR) has drastic performance degradation for class-IL on CIFAR-100, compared to BIR(w/ IR) (\citep[Fig.~8c]{brain_inspired_replay1}).
The used metrics are as follows: (1) Test accuracy (initial accuracy, final accuracy, retention ratio, forgetting score) for all tasks with retention ratio=final accuracy/initial accuracy, forgetting score=initial accuracy - final accuracy (2) Log-likelihood and reconstruction error on the test set (3) Silhouette score representing separability of the cluster. For test accuracy, initial accuracy was measured on the test set right after the model was trained on each particular task, and final accuracy was measured on the test set after the model was sequentially trained on all 10 tasks.


\subsection{Experiment Results}
\subsubsection{Test accuracy comparison throughout all tasks}
\begin{figure}[h]
\vspace{1em}
\centering
\begin{subfigure}[b]{0.48\textwidth}
\includegraphics[width=\linewidth]{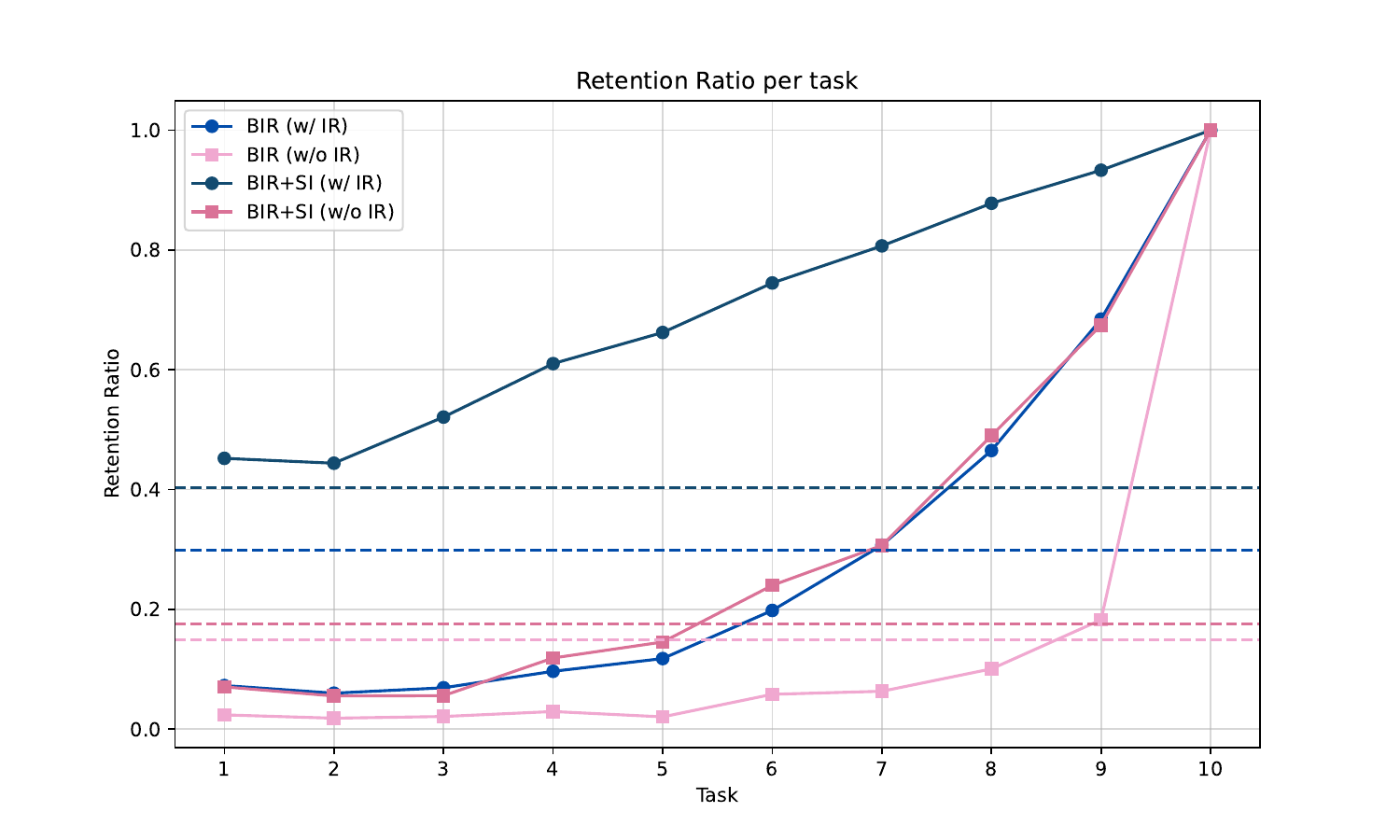}
\end{subfigure}%
\hspace{0.02\textwidth}
\begin{subfigure}[b]{0.48\textwidth}
\includegraphics[width=\linewidth]{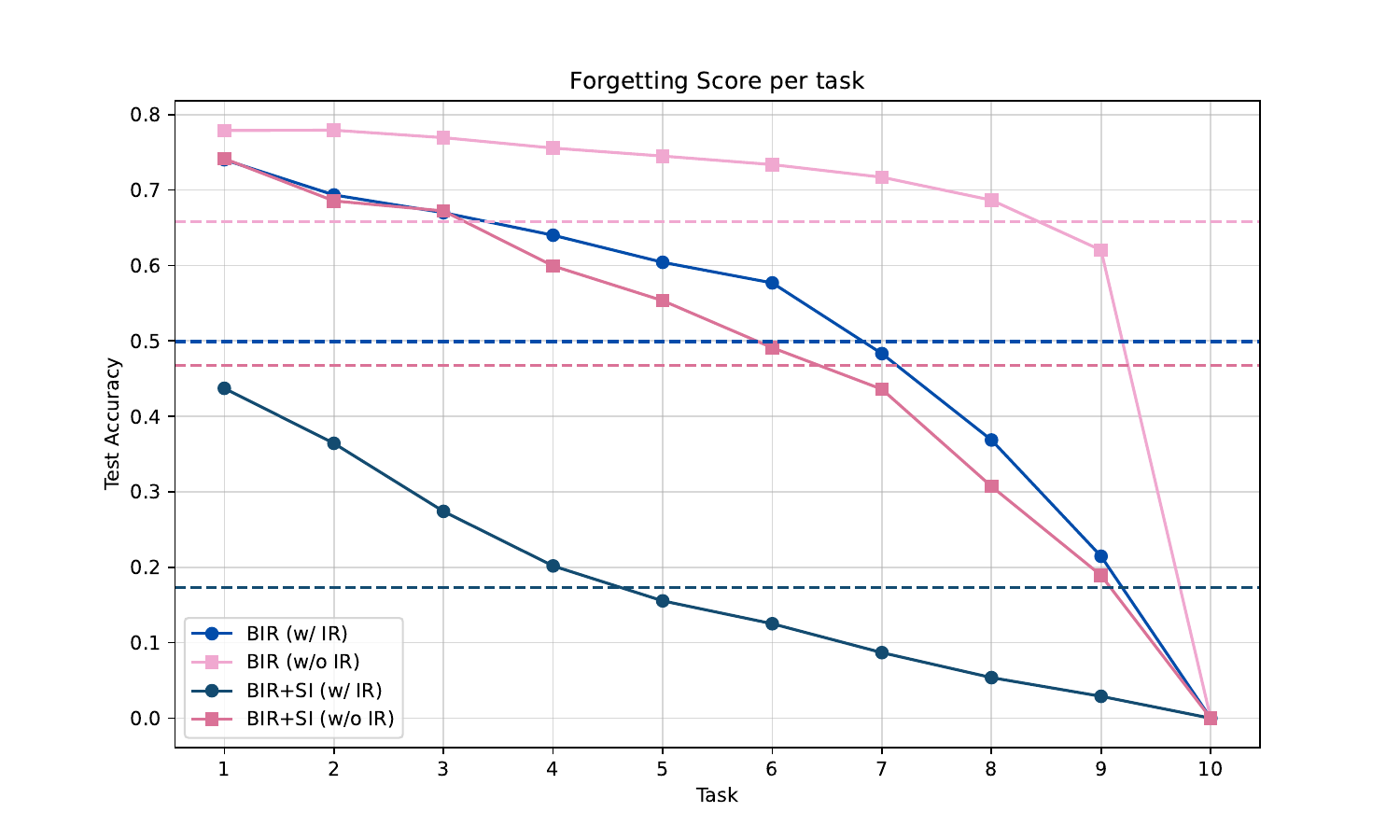}
\end{subfigure}
\caption{
\textbf{Retention ratio and Forgetting score per task.} Retention ratio comparison (\textbf{left}) and forgetting score comparison (\textbf{right}) between BIR(w/ IR), BIR(w/o IR), BIR+SI(w/ IR), BIR+SI(w/o IR) for all tasks. Dashed lines refer to as the average test accuracy throughout all the tasks for each model. 
}
\label{fig:retention_ratio_forgetting_score}
\end{figure}

\begin{figure}[h]
\vspace{1em}
\centering
\begin{subfigure}[b]{0.48\textwidth}
\includegraphics[width=\linewidth]{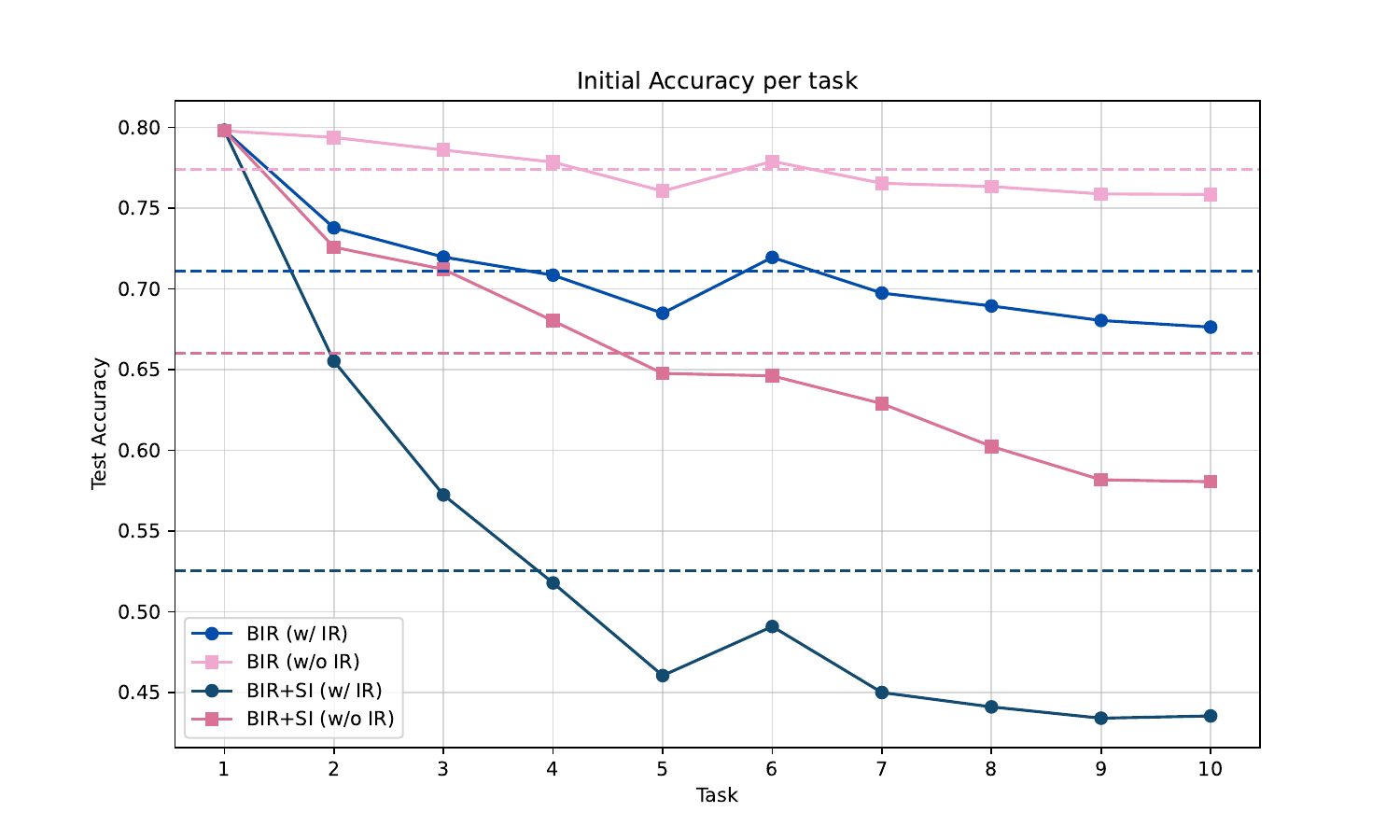}
\end{subfigure}%
\hspace{0.02\textwidth}
\begin{subfigure}[b]{0.48\textwidth}
\includegraphics[width=\linewidth]{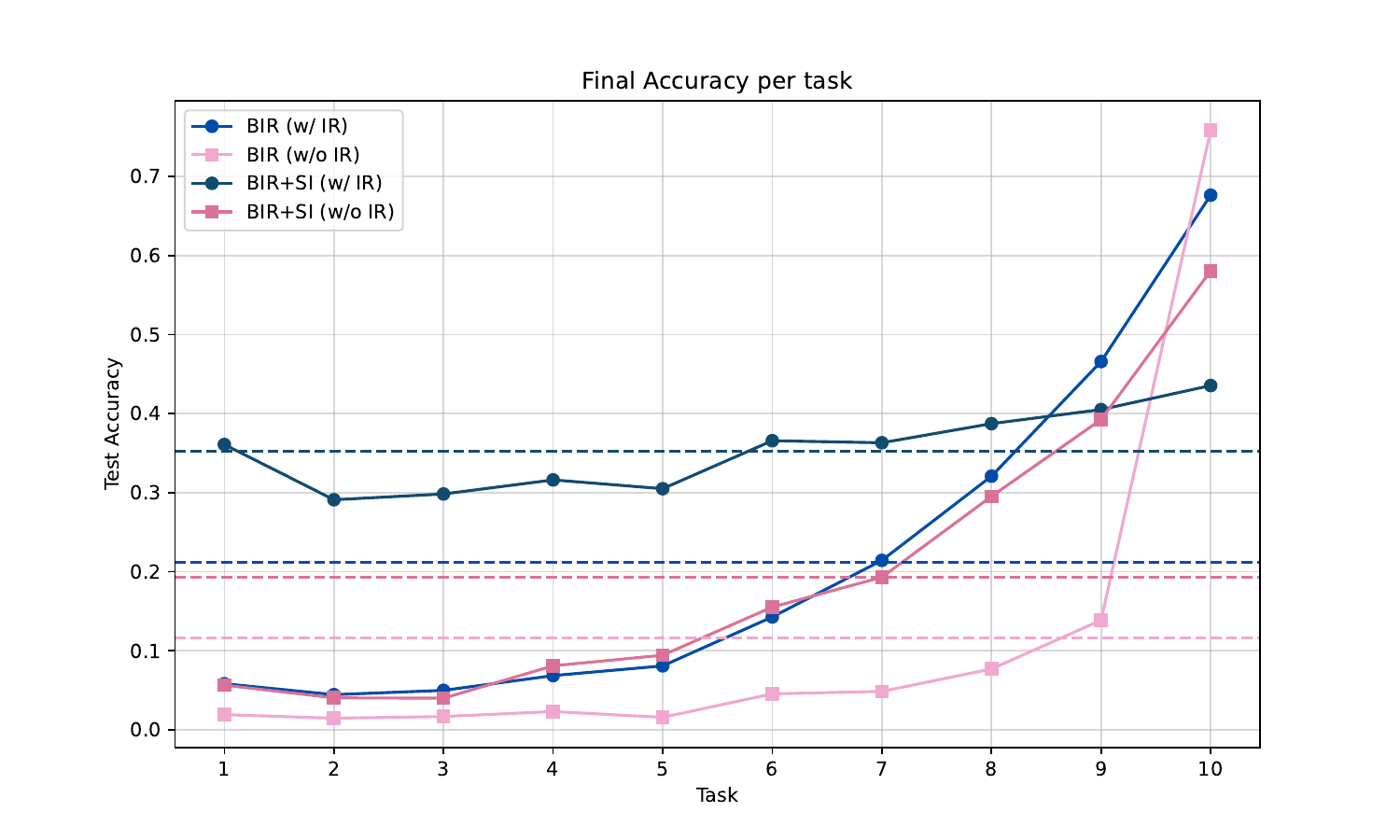}
\end{subfigure}
\caption{
\textbf{Initial accuracy and Final accuracy per task.} Initial test accuracy comparison (\textbf{left}) and final test accuracy comparison (\textbf{right}) between BIR(w/ IR), BIR(w/o IR), BIR + SI(w/ IR), BIR + SI(w/o IR) for all tasks. Dashed lines refer to the average test accuracy throughout all the tasks for each model. 
}
\label{fig:initial_final_accuracy}
\end{figure}

\hspace{0.5cm} As shown in \ref{fig:retention_ratio_forgetting_score}, the BIR+SI (w/ IR) model achieves the highest retention ratio and lowest forgetting score across all tasks. Although both BIR and BIR+SI models perform better in internal replay (w/ IR) conditions than without internal replay (w/o IR), all models except BIR+SI (w/ IR) show drastic catastrophic forgetting across the tasks. Considering that SI(\cite{SI}) protects parameters critical to previous tasks by penalizing large updates to them, the combination of BIR+SI could provide a stable foundation for BIR to replay accurate internal representations. This could be crucial for BIR if the network's parameters drift too much while learning new tasks since BIR depends on the accurate generation of internal representations. 
On the other hand, in \ref{fig:initial_final_accuracy}(left), it is interesting to see that the initial accuracy trend shows the exact opposite of the final accuracy, where the BIR+SI (w/ IR) performs the worst while the BIR (w/o IR) performs the best. As both w/ IR models perform worse than w/o IR models, this shows that internal replay leads to performance degradation when trained on the task, thereby lowering the upper bound of the model's capable performance. Also, since both of the BIR+SI models achieve lower performance than BIR, this shows that the protection made by SI(\cite{SI}) harms the training process, and its contribution comes solely from reducing the catastrophic forgetting of the model. This interpretation also matches with \ref{fig:initial_final_accuracy}(right), as the BIR+SI(w/ IR) shows stable performance in the initial tasks while performing the worst on the last task. The results demonstrate that the proposed models (BIR, BIR+SI) in \cite{brain_inspired_replay1} are not yet optimal, and there could be a better way to reduce the forgetting score while also maintaining the highly achieved initial accuracy in models without internal replay (w/o IR). 

\hspace{0.5cm} 

\subsubsection{Log likelihood distribution and reconstruction error distribution}
\begin{figure}[h]
\vspace{1em}
\centering
\includegraphics[width=\linewidth]{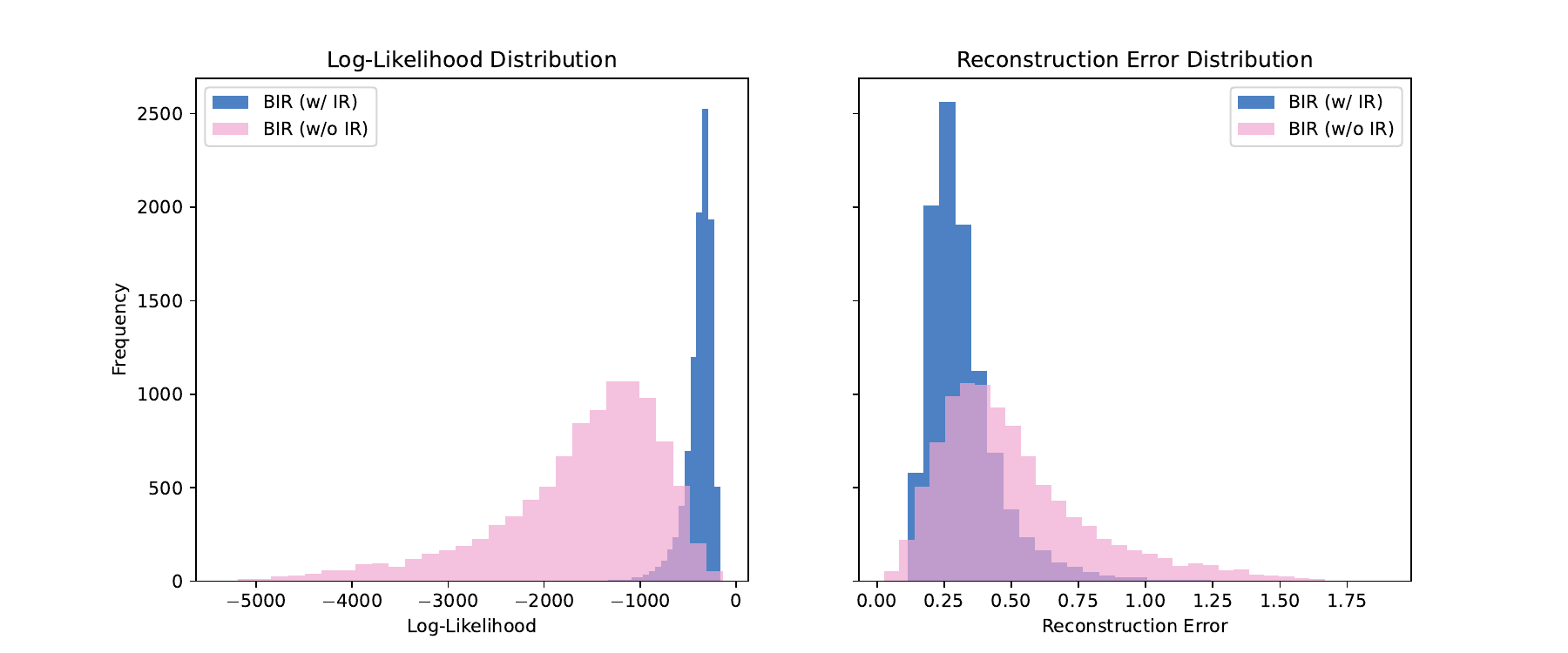}
\caption{
\textbf{Log likelihood distribution and Reconstruction error distribution.} Comparison of log likelihood distribution (\textbf{left}) and reconstruction error distribution (\textbf{right}) between model with internal replay (BIR(w/ IR)) and model without internal replay (BIR(w/o IR)).
}
\label{fig:log_likelihood_reconstruction_error}
\end{figure}

\hspace{0.5cm} Here we present the result of log-likelihood distribution and reconstruction error on the test set. We only provide distributions of BIR models(w/ IR, w/o IR), since the models with BIR+SI showed consistent results with BIR models. In \ref{fig:log_likelihood_reconstruction_error}, BIR(w/ IR) shows higher log-likelihood values, indicating a better fit to the data compared to BIR(w/o IR), which has a longer tail toward more negative values. Moreover, the reconstruction error distribution shows that BIR(w/ IR) achieves consistently lower errors, while BIR(w/o IR) exhibits a wider range of higher reconstruction errors.
\hspace{0.5cm} 

\subsubsection{Silhouette score comparison and UMAP visualization of embedding}
\begin{figure}[h]
\vspace{1em}
\centering
\begin{subfigure}[b]{0.45\textwidth} 
    \includegraphics[width=\linewidth]{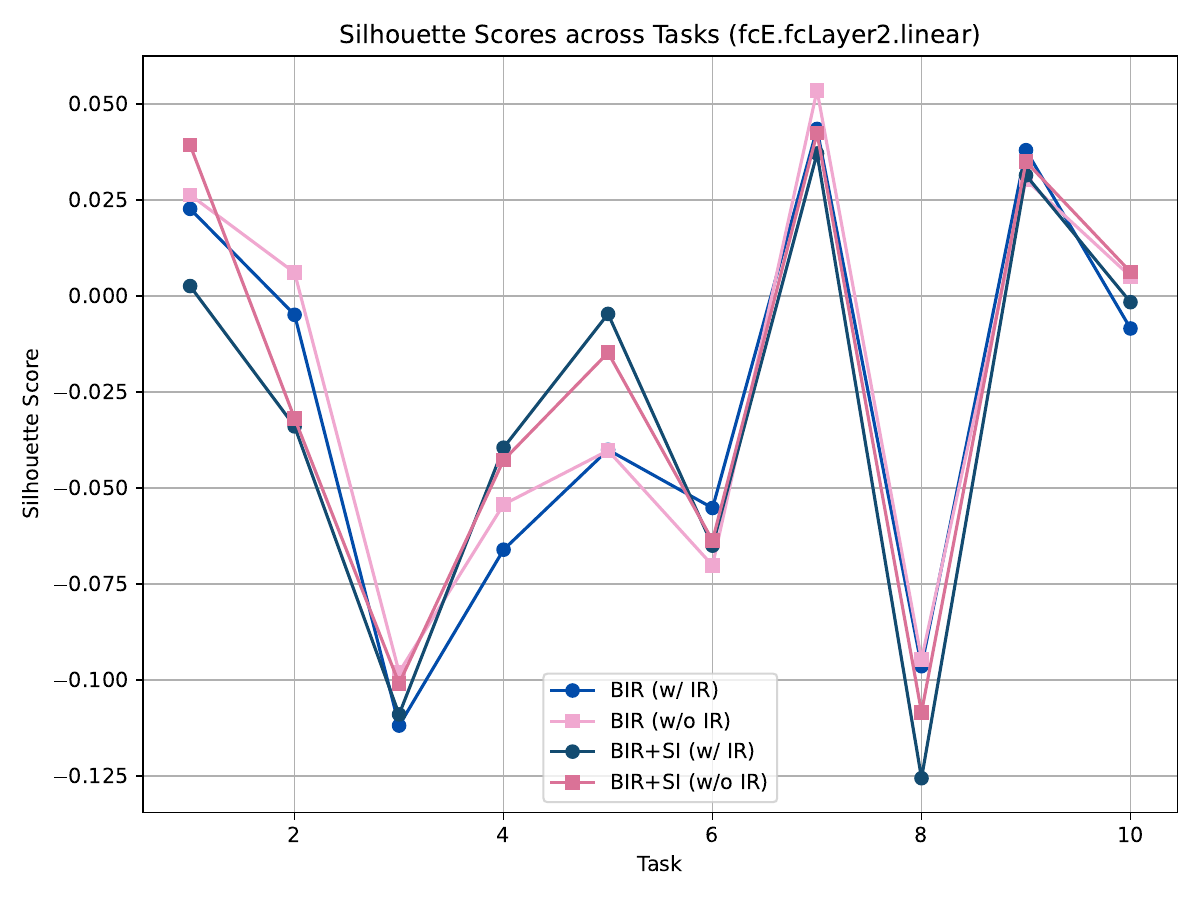}
    \caption{Silhouette Score Comparison}
    \label{fig:left_figure}
\end{subfigure}%
\hspace{0.02\textwidth} 
\begin{minipage}[b]{0.48\textwidth} 
    \centering
    \begin{subfigure}[b]{0.48\textwidth}
        \includegraphics[width=\linewidth]{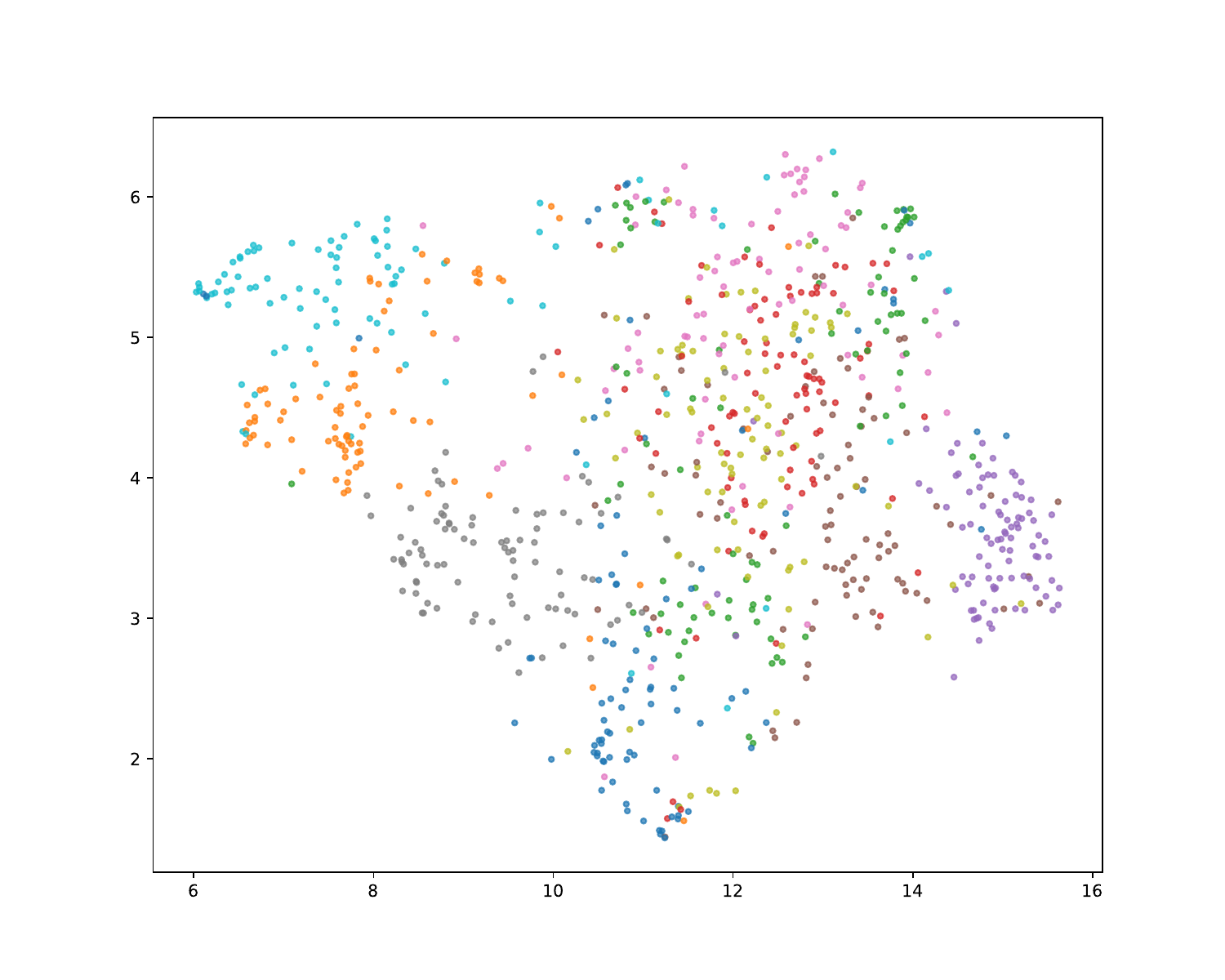}
        \caption{BIR (w/ IR)}
        \label{fig:top_left}
    \end{subfigure}%
    \hspace{0.02\textwidth} 
    \begin{subfigure}[b]{0.48\textwidth}
        \includegraphics[width=\linewidth]{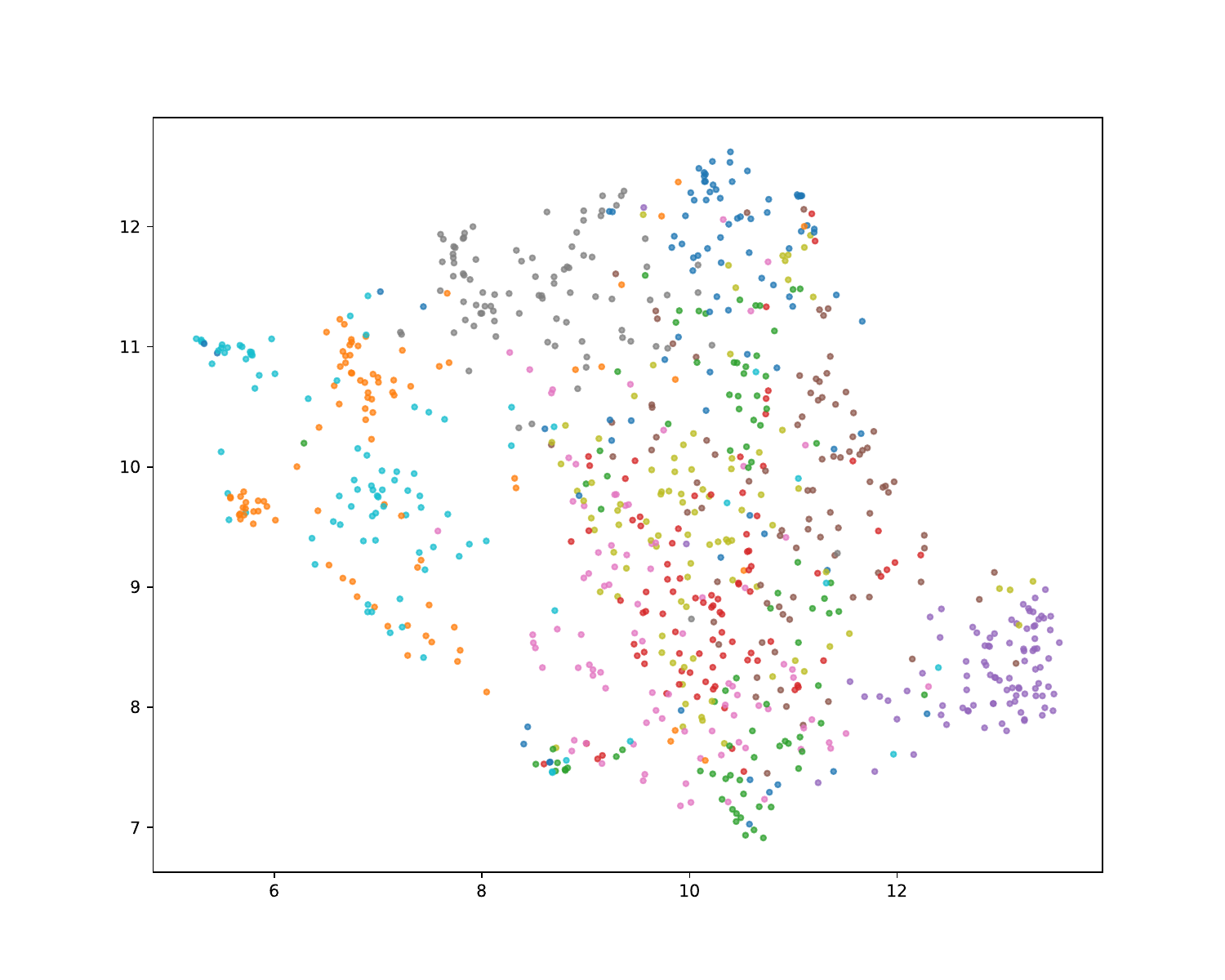}
        \caption{BIR+SI (w/ IR)}
        \label{fig:top_right}
    \end{subfigure}\\[1em]
    \begin{subfigure}[b]{0.48\textwidth}
        \includegraphics[width=\linewidth]{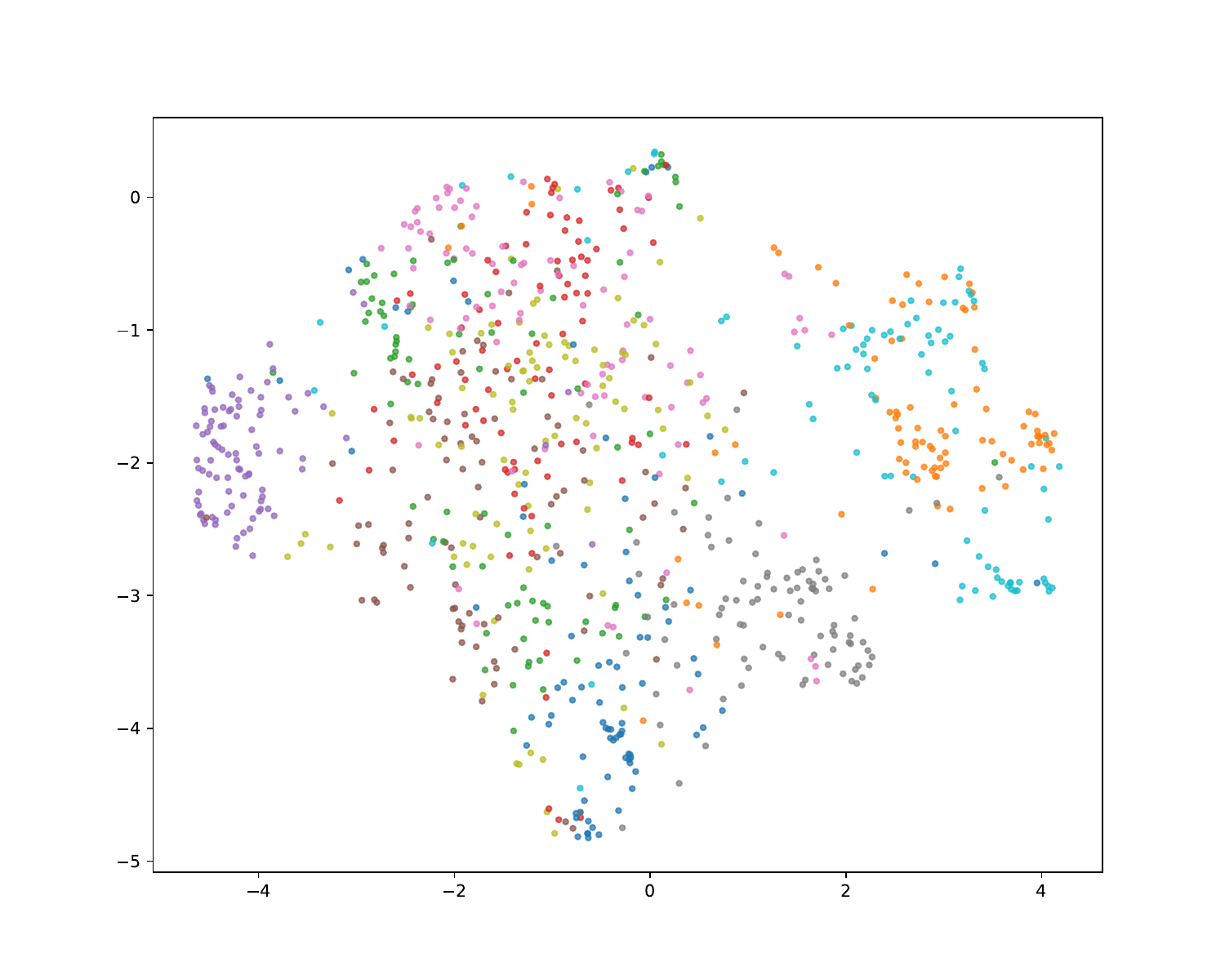}
        \caption{BIR (w/o IR)}
        \label{fig:bottom_left}
    \end{subfigure}%
    \hspace{0.02\textwidth} 
    \begin{subfigure}[b]{0.48\textwidth}
        \includegraphics[width=\linewidth]{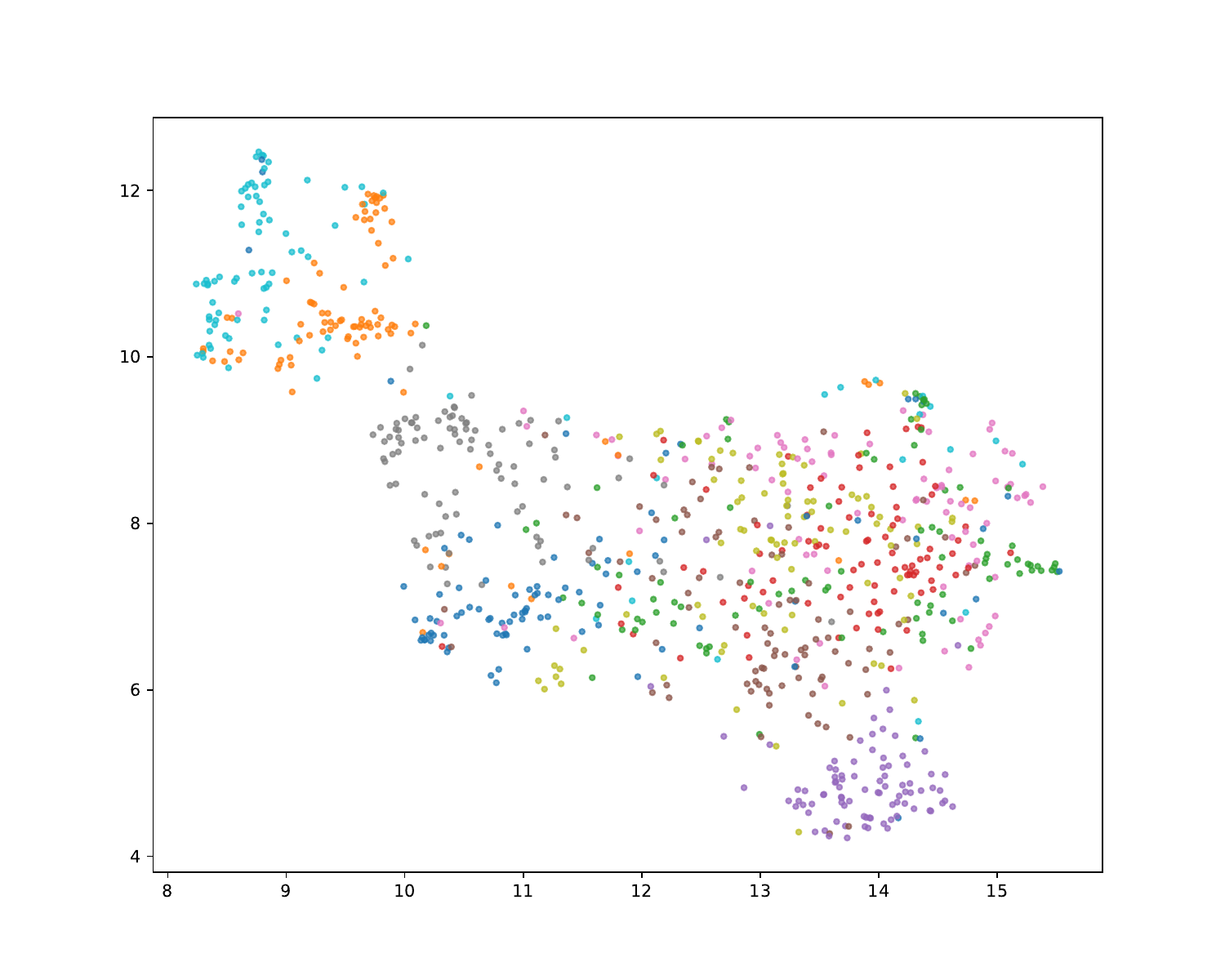}
        \caption{BIR+SI (w/o IR)}
        \label{fig:bottom_right}
    \end{subfigure}
\end{minipage}
\caption{
\textbf{Silhouette score and UMAP visualization of embeddings on layer fcE.fcLayer2.linear.} 
(\textbf{a}) Comparison of silhouette score between model with internal replay (BIR(w/ IR)) and model without internal replay (BIR(w/o IR)). UMAP visualization of embeddings on task 7 for (\textbf{b}) BIR (w/ IR), (\textbf{c}) BIR+SI (w/ IR), (\textbf{d}) BIR (w/o IR), (\textbf{e}) BIR+SI (w/o IR). The experiments are held on layer fcE.fcLayer2.linear.}
\label{fig:silhouette_umap}
\end{figure}

\hspace{0.5cm} To see how the internal replay influences the hidden layer activations over sequential tasks, we extract the hidden layer activations on fcE.fcLayer2.linear for each model using the test set and measure the quality of the clustering using silhouette score for each task. Interestingly, in \ref{fig:silhouette_umap}(a), all models showed very low silhouette score(near 0) with the same tendency across the tasks, indicating poorly separated clusters. In the UMAP visualization result on task 7(\ref{fig:silhouette_umap}(b, c, d, e)), which gave the highest silhouette score compared to other tasks, all models show overlapping in the clusters which means that the hidden layer activations fail to distinctly encode task-specific representations. This demonstrates that the proposed method of \cite{brain_inspired_replay1} is not yet optimal which also aligns with the fact that the final accuracy is not that high(\ref{fig:initial_final_accuracy}(right)).
\hspace{0.5cm}

\section{Conclusion}
\hspace{0.5cm} In this report, we investigate the impact of the brain-inspired replay proposed by \cite{brain_inspired_replay1} on model performance and its ability to address the catastrophic forgetting problem. Our analysis primarily focuses on internal replay, identified as the most critical component of the model. The results show that combining SI (\cite{SI}) with the BIR model, as well as the internal replay itself, helps mitigate catastrophic forgetting (\ref{fig:retention_ratio_forgetting_score}, \ref{fig:initial_final_accuracy}(right)) and improves the model's ability to fit the data (\ref{fig:log_likelihood_reconstruction_error}). However, this combination also led to performance degradation during the training process (\ref{fig:initial_final_accuracy}(right)). Moreover, \ref{fig:silhouette_umap} shows that all the proposed models, regardless of whether they include internal replay, exhibit high overlap in cluster representations. These results demonstrate that while the proposed brain-inspired concept aids in maintaining AI memory, there is room for improvement in addressing the drastic performance degradation observed during training. Therefore, further research should focus not only on resolving catastrophic forgetting but also on maintaining initial accuracy without performance drops.

Additionally, while this report primarily explores the presence or absence of internal replay, investigating the sparsity of the mask for 'gating based on internal context' and the location of the layer where internal replay occurs could provide valuable insights into the model's mechanism and performance. Furthermore, interpreting and extending the model structure using brain mechanisms, as introduced in \cite{neuroscience_textbook}, presents an intriguing avenue for research. As discussed in \ref{appendix:background}, the memory functions of the hippocampus could offer insights into understanding the generative model in brain-inspired GR, as the generative model in the current structure is conceptually aligned with the hippocampus.

For instance, the hippocampus binds sensory information for memory consolidation. By analogizing sensory information to different classes, it would be interesting to explore how the number of classes selected for replay per task impacts performance. Additionally, the hippocampus supports spatial memory, such as the location of behaviorally significant objects, which could inform the 'conditional replay' mechanism of the model. Here, the latent vector of each class could be treated as a form of spatial memory.

Finally, given that both the standard model of memory consolidation and the multiple trace model explain memory consolidation through interactions between the neocortex and hippocampus, it would be fascinating to investigate how these theories align with the models proposed in \cite{brain_inspired_replay1, brain_inspired_replay2, brain_inspired_replay3}.

\bibliographystyle{plainnat}

\clearpage
\bibliography{reference}

\begin{thebibliography}{14}
\providecommand{\natexlab}[1]{#1}
\providecommand{\url}[1]{\texttt{#1}}
\expandafter\ifx\csname urlstyle\endcsname\relax
  \providecommand{\doi}[1]{doi: #1}\else
  \providecommand{\doi}{doi: \begingroup \urlstyle{rm}\Url}\fi

\bibitem[Bear et~al.(2020)Bear, Connors, and Paradiso]{neuroscience_textbook}
Mark Bear, Barry Connors, and Michael~A Paradiso.
\newblock \emph{Neuroscience: exploring the brain, enhanced edition: exploring the brain}.
\newblock Jones \& Bartlett Learning, 2020.

\bibitem[Carr et~al.(2011)Carr, Jadhav, and Frank]{brain_mem_sleep}
Margaret~F Carr, Shantanu~P Jadhav, and Loren~M Frank.
\newblock Hippocampal replay in the awake state: a potential substrate for memory consolidation and retrieval.
\newblock \emph{Nature neuroscience}, 14\penalty0 (2):\penalty0 147--153, 2011.

\bibitem[Ji and Wilson(2007)]{brain_mem2}
Daoyun Ji and Matthew~A Wilson.
\newblock Coordinated memory replay in the visual cortex and hippocampus during sleep.
\newblock \emph{Nature neuroscience}, 10\penalty0 (1):\penalty0 100--107, 2007.

\bibitem[Millichamp and Chen(2021)]{brain_inspired_replay2}
Jack Millichamp and Xi~Chen.
\newblock Brain-inspired feature exaggeration in generative replay for continual learning.
\newblock \emph{arXiv preprint arXiv:2110.15056}, 2021.

\bibitem[Oudiette and Paller(2013)]{brain_memory_3}
Delphine Oudiette and Ken~A Paller.
\newblock Upgrading the sleeping brain with targeted memory reactivation.
\newblock \emph{Trends in cognitive sciences}, 17\penalty0 (3):\penalty0 142--149, 2013.

\bibitem[Qin et~al.(1997)Qin, Mcnaughton, Skaggs, and Barnes]{brain_mem1}
Yu-Lin Qin, Bruce~L Mcnaughton, William~E Skaggs, and Carol~A Barnes.
\newblock Memory reprocessing in corticocortical and hippocampocortical neuronal ensembles.
\newblock \emph{Philosophical Transactions of the Royal Society of London. Series B: Biological Sciences}, 352\penalty0 (1360):\penalty0 1525--1533, 1997.

\bibitem[Ran et~al.(2024)Ran, Yao, Wang, Xu, and Liu]{brain_inspired_replay3}
Xuming Ran, Juntao Yao, Yusong Wang, Mingkun Xu, and Dianbo Liu.
\newblock Brain-inspired continual pre-trained learner via silent synaptic consolidation.
\newblock \emph{arXiv preprint arXiv:2410.05899}, 2024.

\bibitem[Rasch and Born(2007)]{brain_memory_2}
Bj{\"o}rn Rasch and Jan Born.
\newblock Maintaining memories by reactivation.
\newblock \emph{Current opinion in neurobiology}, 17\penalty0 (6):\penalty0 698--703, 2007.

\bibitem[Scoville and Milner(1957)]{brain_sleep_prev1}
William~Beecher Scoville and Brenda Milner.
\newblock Loss of recent memory after bilateral hippocampal lesions.
\newblock \emph{Journal of neurology, neurosurgery, and psychiatry}, 20\penalty0 (1):\penalty0 11, 1957.

\bibitem[Squire(1992)]{brain_sleep_prev2}
Larry~R Squire.
\newblock Memory and the hippocampus: a synthesis from findings with rats, monkeys, and humans.
\newblock \emph{Psychological review}, 99\penalty0 (2):\penalty0 195, 1992.

\bibitem[Van~de Ven et~al.(2016)Van~de Ven, Trouche, McNamara, Allen, and Dupret]{brain_memory_4}
Gido~M Van~de Ven, Stephanie Trouche, Colin~G McNamara, Kevin Allen, and David Dupret.
\newblock Hippocampal offline reactivation consolidates recently formed cell assembly patterns during sharp wave-ripples.
\newblock \emph{Neuron}, 92\penalty0 (5):\penalty0 968--974, 2016.

\bibitem[Van~de Ven et~al.(2020)Van~de Ven, Siegelmann, and Tolias]{brain_inspired_replay1}
Gido~M Van~de Ven, Hava~T Siegelmann, and Andreas~S Tolias.
\newblock Brain-inspired replay for continual learning with artificial neural networks.
\newblock \emph{Nature communications}, 11\penalty0 (1):\penalty0 4069, 2020.

\bibitem[Wilson and McNaughton(1994)]{brain_memory_1}
Matthew~A Wilson and Bruce~L McNaughton.
\newblock Reactivation of hippocampal ensemble memories during sleep.
\newblock \emph{Science}, 265\penalty0 (5172):\penalty0 676--679, 1994.

\bibitem[Zenke et~al.(2017)Zenke, Poole, and Ganguli]{SI}
Friedemann Zenke, Ben Poole, and Surya Ganguli.
\newblock Continual learning through synaptic intelligence, 2017.
\newblock URL \url{https://arxiv.org/abs/1703.04200}.

\end{thebibliography}

\clearpage
\appendix

\section{Memory system in human brain}
\label{appendix:background}
\hspace{0.5cm} According to \citep[p.~838]{neuroscience_textbook}, the temporal lobe is the important brain region for recording past events. The temporal neocortex in the medial temporal lobe may be a site of long-term memory storage, and the medial temporal lobe also contains a group of structures interconnected with the neocortex that are critical for forming declarative 
memories. Also, \citep[p.~846]{neuroscience_textbook} summarizes the memory function of the hippocampus, which has received the greatest attention among the medial temporal lobe region. First of all, it plays a critical role in binding sensory information for memory consolidation. Second, the hippocampus supports spatial memory of the location of objects of behavioral importance. Lastly, the hippocampus is involved in the storage of memories for some length of time. Moreover, \citep[Figure 24.25, p.~855]{neuroscience_textbook} illustrates two popular hypotheses of memory consolidation (i.e. standard model of memory consolidation, multiple trace model of consolidation) and both explain the memory consolidation process as the interaction between neocortex and hippocampus. The standard model of memory consolidation suggests that during synaptic consolidation, memory retrieval requires the hippocampus but after it is complete, the systems consolidation occurs in which engrams are moved gradually into distributed areas of the neocortex. On the other hand, the multiple trace model of consolidation proposes the idea that each time an episodic memory is retrieved, it occurs in a context different from the initial experience and the recalled information combines with new sensory input to form a new memory trace involving both the hippocampus and neocortex.

Previous works such as \cite{brain_sleep_prev1, brain_sleep_prev2} have led to the general hypothesis that the hippocampus could be facilitating the reactivation of neocortical activity patterns during 'offline' periods such as sleep or quiet awake when the neocortex is not actively engaged in processing incoming data. This reactivation has been presumed to permit recently acquired information to become appropriately integrated into long-term memory. On top of that, \cite{brain_mem1, brain_mem2, brain_mem_sleep} have shown that memory replay is orchestrated by the hippocampus but also observed in the cortex, and mainly occurs in the sharp-wave/ripples during both sleep and awake. Inspired by the process of memory consolidation in the brain, \cite{brain_inspired_replay1, brain_inspired_replay2, brain_inspired_replay3} have explored the potential of leveraging this biological mechanism in ANNs to address the issue of catastrophic forgetting in continual learning.

\section{Detailed explanation of the model structure}
\label{appendix:brain_replay_description}
\hspace{0.5cm} Here we provide a detailed explanation of each component of the model in \cite{brain_inspired_replay1}.
Inspired by the brain anatomy in which the hippocampus sits atop the cortex in the brain processing hierarchy, the replay through feedback merges the generator(hippocampus) into the main model(cortex) and uses a single variational autoencoder(VAE) model with additional softmax classification layer to operate as the generator and the main model. Therefore, in this model, the early layers are considered the visual cortex, and the top layers are considered the hippocampus. Next, since humans can have control over what memories to recall while VAE doesn’t, the standard normal prior over the VAE’s latent variable is replaced by the Gaussian mixture with a separate mode for each class. This enables the model to generate specific classes by restricting the sampling of the latent variables to their corresponding modes. The third component is gating based on internal context, which is inspired by the brain processing stimuli differently depending on the context or task that is performed and also that the contextual cues bias what memories are replayed. Therefore, a context gate that randomly selects a subset of neurons is used in each hidden layer depending on which task should be performed. Additionally, considering that the brain doesn’t replay memories down to the input level, for example as the mental images are not propagated to our retina, the internal replay replays the representations of previously learned classes at the hidden level, rather than at the input level (e.g. pixel level). Finally, unlike the straightforward GR that labels the generated data into one specific class, brain-inspired GR uses soft targets that are labeled with predicted probabilities for all possible classes, which it refers to as distillation. Such an approach is especially important when the quality of the generated data is low so when it is hard to classify as a single class.
\end{document}